%% file: paper_with_supplemental.tex
\documentclass[runningheads]{llncs}

\usepackage{eccv}

\usepackage{eccvabbrv}

\usepackage{graphicx}
\usepackage{booktabs}

\usepackage[accsupp]{axessibility}  % Improves PDF readability for those with disabilities.
\usepackage{rotating}

\usepackage{hyperref}

\usepackage{orcidlink}
\usepackage{listings}

\definecolor{mauve}{rgb}{0.855, 0.65, 0.1255}
\begin{document}

\title{2S-ODIS: Two-Stage Omni-Directional Image Synthesis by Geometric Distortion Correction}

\titlerunning{2S-ODIS: Two-Stage Omni-Directional Image Synthesis}

%\author{First Author\inst{1}\orcidlink{0000-1111-2222-3333} \and
%Second Author\inst{2,3}\orcidlink{1111-2222-3333-4444} \and
%Third Author\inst{3}\orcidlink{2222--3333-4444-5555}}

\author{Atsuya Nakata\orcidlink{0009-0002-7815-9604} \and
Takao Yamanaka\orcidlink{0000-0001-9028-8244}}
%\inst{1}
\authorrunning{A. Nakata and T. Yamanaka}

\institute{Sophia University, Tokyo, Japan \\
\email{a-nakata-7r0@eagle.sophia.ac.jp, takao-y@sophia.ac.jp}}

%\authorrunning{F.~Author et al.}
%% First names are abbreviated in the running head.
%% If there are more than two authors, 'et al.' is used.
%
%% TODO FINAL: Replace with your institution list.
%\institute{Princeton University, Princeton NJ 08544, USA \and
%Springer Heidelberg, Tiergartenstr.~17, 69121 Heidelberg, Germany

\maketitle

\input{sec/0_abstract}
\input{sec/1_intro}
\input{sec/2_relatedworks}
\input{sec/3_proposedmethod}
\input{sec/4_experiment}
\input{sec/5_result}
\input{sec/6_limit}
\input{sec/7_conc}

\section*{Acknowledgement}
This work was supported by JSPS KAKENHI Grant Number JP21K11943

%\clearpage  % TODO REVIEW/FINAL: This \clearpage needs to be removed from both review and camera-ready versions.

\bibliographystyle{splncs04}
\bibliography{main}

\clearpage 
\appendix
\section*{Supplemental}
\setcounter{figure}{9}
\setcounter{table}{3}
%\section{Appendix}
\input{sec/X_suppl}

%\setcounter{enumiv}{20}
%\bibliography{main_appendix}

% \begin{thebibliography}{99}
% \bibitem[21]{multiaxistransformer} Zhao, L., Zhang, Z., Chen, T., Metaxas, D., Zhang, H.: Improved transformer for high-resolution gans. In: Advances in Neural Information Processing Systems(NeurIPS). vol. 34, pp. 18367–18380 (2021)
% \end{thebibliography}

\end{document}

%% file: sec/0_abstract.tex
\begin{abstract}
Omni-directional images have been increasingly used in various applications, including virtual reality
and SNS (Social Networking Services).
However, their availability is comparatively limited in contrast to normal field of view (NFoV) images,
since specialized cameras are required to take omni-directional images.
Consequently, several methods have been proposed based on generative adversarial networks (GAN) to synthesize omni-directional images,
but these approaches have shown difficulties in training of the models,
due to instability and/or significant time consumption in the training.
To address these problems, this paper proposes a novel omni-directional image synthesis method,
2S-ODIS (Two-Stage Omni-Directional Image Synthesis), which generated high-quality omni-directional images
but drastically reduced the training time.
This was realized by utilizing the VQGAN (Vector Quantized GAN) model pre-trained on a large-scale NFoV image database
such as ImageNet without fine-tuning.
Since this pre-trained model does not represent distortions of omni-directional images in the equi-rectangular projection (ERP),
it cannot be applied directly to the omni-directional image synthesis in ERP.
Therefore, two-stage structure was adopted to first create a global coarse image in ERP and then refine the image
by integrating multiple local NFoV images in the higher resolution to compensate the distortions in ERP,
both of which are based on the pre-trained VQGAN model.
As a result, the proposed method, 2S-ODIS, achieved the reduction of the training time from 14 days in
OmniDreamer to four days in higher image quality.
\end{abstract}

%% file: sec/1_intro.tex
\begin{figure}
    \centering
    \includegraphics[width=\linewidth]{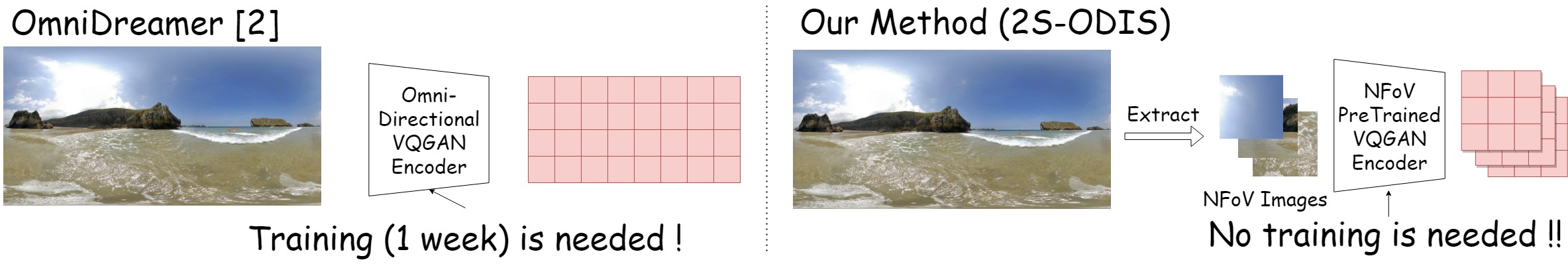}
    \caption{
        Overview of advantages of proposed method, 2S-ODIS.
        OmniDreamer~\cite{omnidreamer} requires 14 days for training of the model,
        including 1-week training of the VQGAN model.
        In contrast, the proposed method only required 4 days for the training of the model since no training of VQGAN
        model was required.
    }
    \label{fig:abst_method}
\end{figure}

\section{Introduction}
\label{sec:intro}

An omni-directional image is taken by an omni-directional camera, also known as a 360-degree camera,
which captures lights from all directions falling onto the focal point to cover a full sphere,
and is usually represented in the equi-rectangular projection (ERP) to represent it in a 2-dimensional plane.
These images have been applied to various applications such as virtual reality, social networking services,
and map tools such as Google Street View.
However, the availability of the omni-directional images are still limited compared with
Normal Field of View (NFoV) images captured by a standard camera, since the specialized camera is required to
capture the omni-directional images.

Recently, several methods have been proposed to synthesize omni-directional images from a NFoV
image~\cite{sphericalgan, odimlpmixergan, omnidreamer, GuidedImmerseGAN, SphericalGenerationSymmetry},
a text condition~\cite{text2light}, or both~\cite{aognet}.
These methods have been based on Generative Adversarial Networks (GAN)~\cite{sphericalgan, odimlpmixergan, GuidedImmerseGAN, SphericalGenerationSymmetry},
Vector Quantized GAN (VQGAN)~\cite{omnidreamer,text2light}, or auto-regressive outpainting using stable diffusion~\cite{aognet}.
However, the GAN-based methods have faced challenges of instability in training,
while the VQGAN-based methods and the auto-regression with stable diffusion require long training and inference time, respectively.

In this paper, we propose a novel method to synthesize omni-directional images from a NFoV image based on pre-trained VQGAN,
trained on a large-scale NFoV-image dataset.
The previous method with VQGAN~\cite{omnidreamer} has required to train the VQGAN encoder and decoder to represent geometric
distortions in omni-directional images in ERP, especially at poles (top and bottom regions in ERP).
This training has taken long time, such as 1 week in the method~\cite{omnidreamer}.
In the proposed method, the pre-trained VQGAN encoder and decoder were applied without fine-tuning by synthesizing multiple
NFoV images to integrate them into an omni-directional image based on geometric distortion correction.
Since no training of VQGAN was required, the training of the model was shortened by removing the step of the fine-tuning
of VQGAN, as shown in Fig.~\ref{fig:abst_method}.
Furthermore, a two-stage structure was adopted in the proposed method.
At the first stage, a global coarse image in ERP is created using the pre-trained VQGAN encoder and decoder without the
geometric distortion correction.
Therefore, the generated omni-directional image at the first stage includes distortions.
For example, a straight line in NFoV images at poles can not be reproduced at the first stage.
At the second stage,
this global coarse image is refined by synthesizing an omni-directional image from multiple NFoV images generated
using the pre-trained VQGAN encoder and decoder.
This second stage compensates the geometric distortions at the first stage,
in addition to representing local detailed textures in a higher resolution.
By using the two-stage structure,
the model can produce globally plausible yet locally detailed omni-directional images without the geometric distortions.

The contributions of this paper include:
\begin{itemize}
    \item A novel method to synthesize omni-directional images from a NFoV image was proposed using pre-trained VQGAN.
Since no training of VQGAN was required, the training time was drastically reduced.
    \item A two-stage structure was adopted to generate a global coarse omni-directional image at the first stage,
and then generate a locally detailed image with geometric distortion correction at the second stage.
    \item Experimental results demonstrated that the proposed method synthesized higher quality omni-directional
images in shortened training and inference time than the previous methods such as OmniDreamer~\cite{omnidreamer}.
\end{itemize}

%% file: sec/2_relatedworks.tex
\section{Related Works}
\label{sec:relatedworks}

\subsection{Image Generation}

VQVAE (Vector Quantized Variational AutoEncoder)~\cite{vqvae} has been proposed to improve the generated image blurriness in
VAE~\cite{vae} by representing image patches with quantized latent vectors based on vector quantization.
Furthermore, the adversarial loss has been introduced in VQVAE to make the generated images clearer, called VQGAN~\cite{vqgan}.
In this method, Transformer~\cite{transformer} has been used to sequentially predict image patches from neighbor patches based on
auto-regressive prediction.
The patches are represented with the quantized latent vectors called VQGAN codes to generate clear images with
low computational cost.
To improve the slow inference in VQGAN due to the sequential predictions of patches,
MaskGIT~\cite{maskgit} has been proposed by predicting multiple patches simultaneously.
Although MaskGIT has succeeded in improving the inference speed, it has been difficult to generate high quality images
in the high resolution.
To solve this problem, Muse~\cite{muse} has been proposed using a two-stage structure,
where a low-resolution image is generated at the first stage, and then is refined to generate a higher-resolution image
at the second stage.
In our proposed method, this two-stage structure was adopted for the omni-directional image synthesis in the high resolution.

\subsection{Omni-directional Image Generation}
Several methods have been proposed for synthesizing omni-directional images from NFoV images.
Okubo and Yamanaka~\cite{sphericalgan} have proposed a method of generating omni-directional images based on conditional GAN
from a single NFoV image with the class label.
Hara et al.~\cite{SphericalGenerationSymmetry} have also proposed a method based on the symmetric property in the omni-directional images using GAN and VAE.
Another work to synthesize omni-directional images is Guided ImmerseGAN\cite{GuidedImmerseGAN}, which generates omni-directional images
from a NFoV image with the modulation guided by a given class label, which does not have to be the true class of the input NFoV image.
In the work of OmniDreamer~\cite{omnidreamer}, VQGAN has been applied to the omni-directional image synthesis by using Transformer for
auto-regressive prediction.
In this method, VQGAN encoder and decoder have to be fine-tuned on an omni-directional image dataset since the geometric
distortion in ERP has to be represented in the latent codes of VQGAN.
Text2Light~\cite{text2light} also uses VQGAN with auto-regressive prediction for the generation,
though only text information is taken as the input instead of the NFoV image.
Nakata et al.~\cite{odimlpmixergan} have proposed a method to increase the diversity of generated omni-directional images
based on MLP-Mixer by efficiently propagating the information of the NFoV image embedded at the center in ERP.
AOGNet~\cite{aognet} has generated omni-directional images by out-painting an incomplete 360-degree image progressively with
NFoV and text guidances jointly or individually.
This has been realized using auto-regressive prediction based on the stable-diffusion backbone model.
Due to the nature of sequential auto-regressive prediction, it takes long inference time.

In our proposed method, the pre-trained VQGAN model was used without fine-tuning on the omni-directional image dataset,
since multiple NFoV images are synthesized based on the pre-trained VQGAN and then integrated into omni-directional images.
By removing the step of VQGAN training, the overall training of the model was drastically shortened than the previous
method with VQGAN such as OmniDreamer~\cite{omnidreamer}.
In addition, the proposed method is based on simultaneous synthesis of multiple NFoV images in different directions,
whose inference was faster than auto-regressive prediction such as OmniDreamer~\cite{omnidreamer} and AOGNet~\cite{aognet}.

%% file: sec/3_proposedmethod.tex
\section{Proposed Method}
\begin{figure*}[t]
    \centering
    \includegraphics[width=\linewidth]{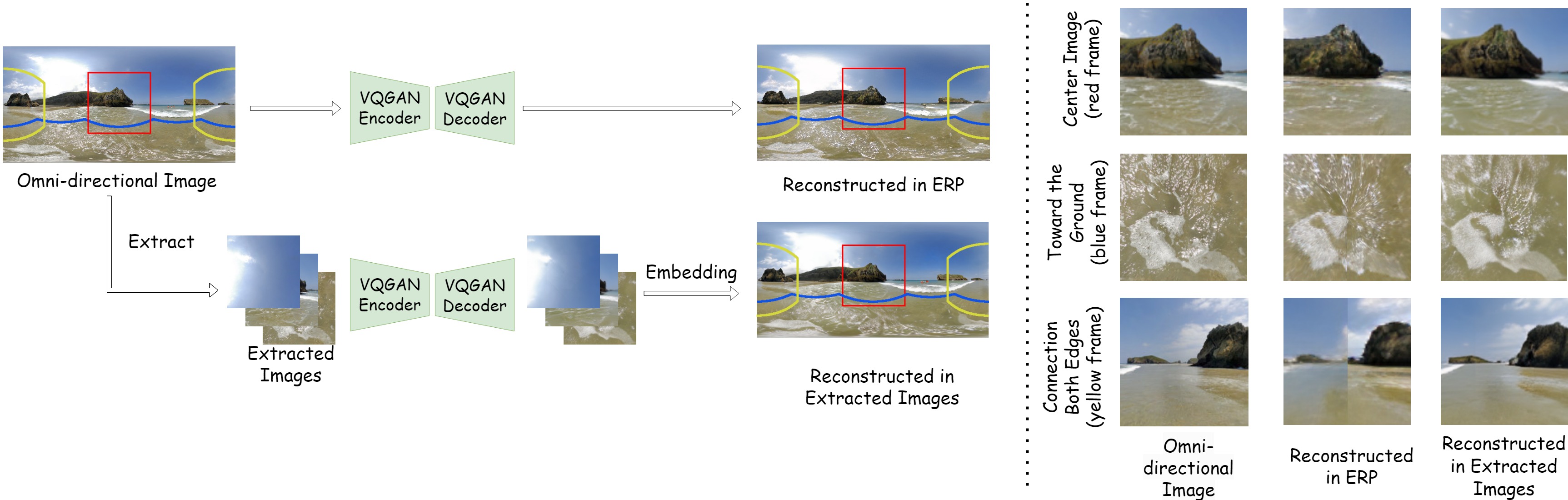}
    \caption{
        Qualitative comparison of omni-directional image reconstruction using pre-trained VQGAN encoder and decoder.
        Omni-directional Image: original omni-directional image,
        Reconstructed in ERP: reconstructed in equirectangular projection,
        Reconstructed in Extracted Images: reconstructed by integrating multiple NFoV images in different directions.
        By extracting NFoV images, an omni-directional image can be correctly reconstructed without distortions.
    }
    \label{fig:reconstruct_difference}
\end{figure*}
\begin{figure*}[t]
    \centering
    \includegraphics[width=\linewidth]{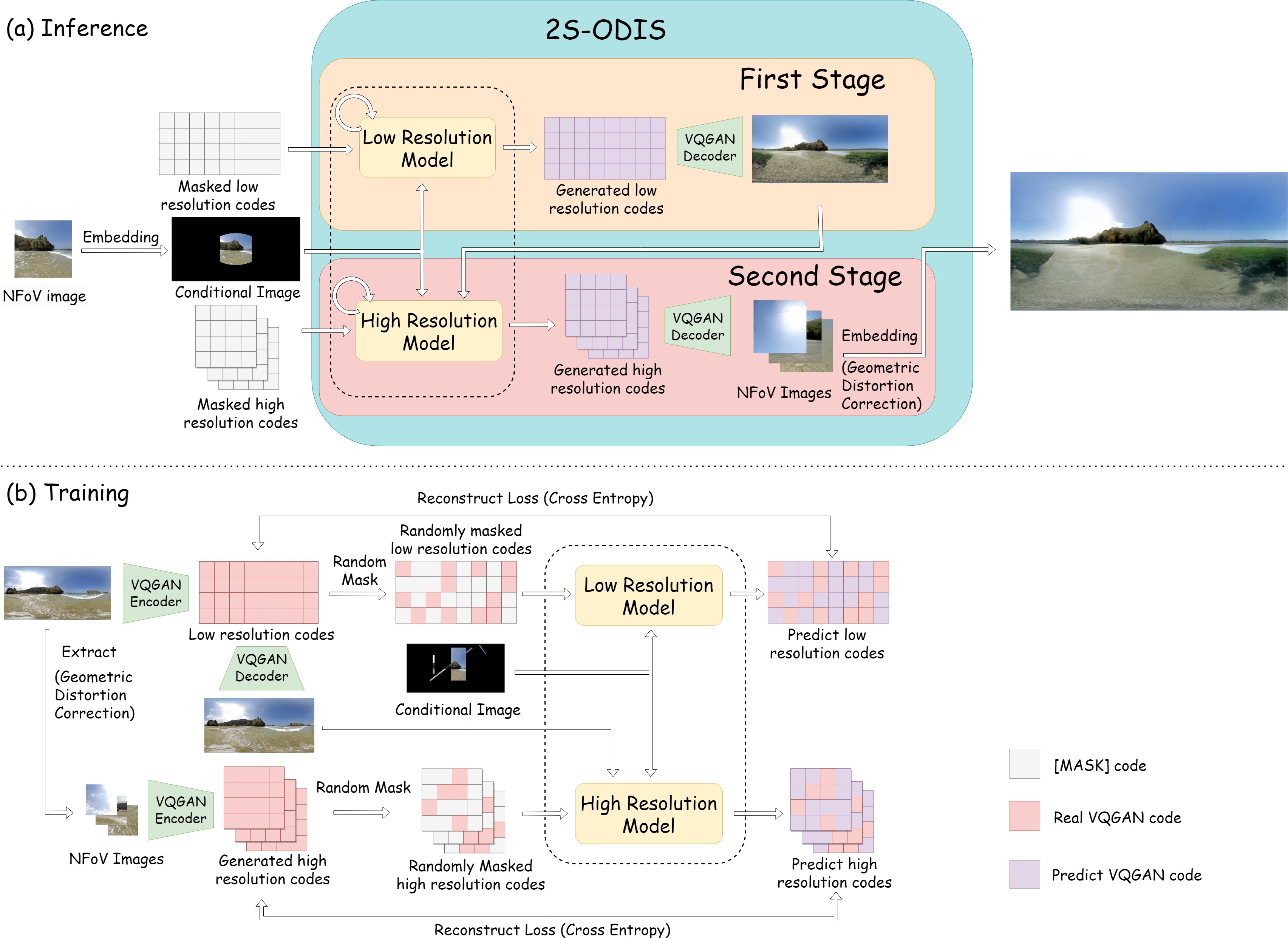}
    \caption{Diagram of the proposed method. (a)Inference, (b)Training.}
    \label{fig:proposed_method}
\end{figure*}
\begin{figure}[t]
    \centering
    \includegraphics[width=\linewidth]{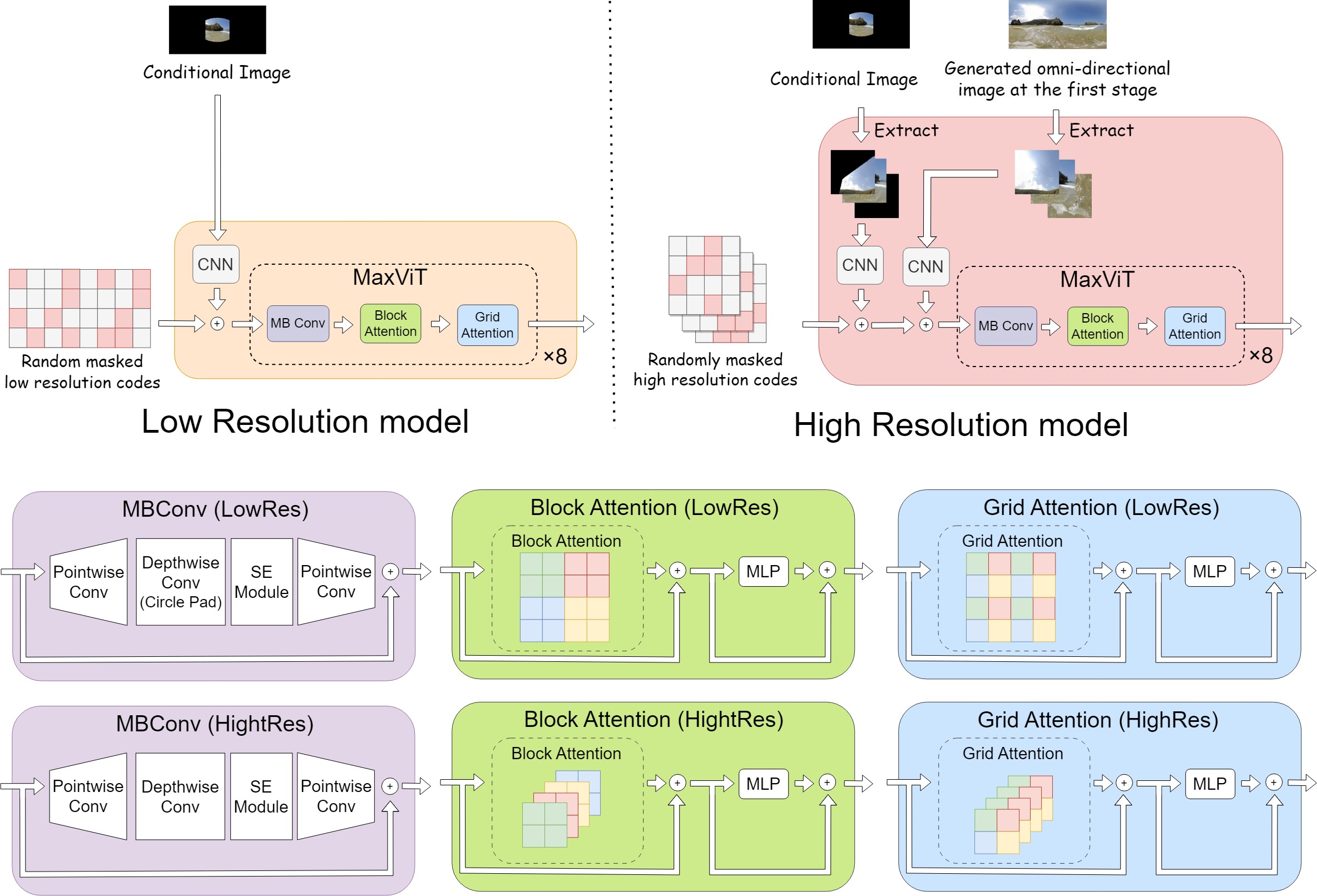}
    \caption{Structure of the proposed method. The range of attention differs between the high-resolution and low-resolution models.}
    \label{fig:attention_architecture}
\end{figure}

\subsection{Two-Stage Structure}
The proposed method consists of the two-stage structure, where a global coarse omni-directional image in ERP is
synthesized in a low resolution (256$\times$512 pixels) at the first stage without geometric distortion correction,
and then is refined at the second stage by integrating the multiple synthesized NFoV images in different directions
based on the geometric distortion correction, producing a high-quality omni-directional image in ERP in a high
resolution (1024$\times$2048 pixels).
At both stages, the pre-trained VQGAN was utilized without fine-tuning on the omni-directional image dataset.

As a preliminary experiment, an omni-directional image was reconstructed in ERP or in multiple NFoV images using
pre-trained VQGAN encoder and decoder without the fine-tuning, as shown in Fig.~\ref{fig:reconstruct_difference}.
It can be seen from the figure that the reconstruction in ERP cannot correctly reproduce the texture in the region
toward the ground (blue frame) and the continuity in the region at both edges (yellow frame),
although it can reproduce the region at center in ERP (red frame).
On the contrary, all the regions can be correctly reproduced in the extracted NFoV images.
This indicates that the pre-trained VQGAN model can be applied without fine-tuning if it is applied to NFoV images.

Thus, the generated omni-directional image in ERP at the first stage in the proposed method includes distortions since
the pre-trained VQGAN cannot represent the texture and continuities in the omni-directional images in ERP.
However, these distortions are correctly compensated at the second stage by synthesizing the multiple NFoV images
which can be correctly reproduced by the pre-trained VQGAN model.
If only the second stage is used in the proposed method, it is difficult to synthesize multiple NFoV images
simultaneously with global compatibility.
Therefore, the two-stage structure was adopted in the proposed method to produce globally plausible coarse
omni-directional image at the first stage.

The structure of the proposed method is shown in Fig.~\ref{fig:proposed_method}.
At the first stage, the low resolution model produces the low resolution codes, which are converted into patches of
omni-directional images in ERP using the pre-trained VQGAN decoder.
At the second stage, the high resolution model produces the high resolution codes, which are corresponding to the
patches in the NFoV images in multiple directions (26 directions in our implementation) in an omni-directional image
with overlapping.
These 26 directions for the NFoV images were the same directions as the normal vectors in the faces of a rhombicuboctahedron.
The field of view was set to 60 degrees for all directions.
The generated NFoV images with the size of 256$\times$256 pixels were integrated into an omni-directional image with the
size of 1024$\times$2048 pixels in ERP.

\subsection{Inference}
For synthesizing an omni-directional image in the inference, the low-resolution codes are first generated using the
sampling strategy proposed in MaskGIT~\cite{maskgit} at the first stage from the conditional image where an input NFoV image is
embedded at the center in ERP, as shown in Fig.~\ref{fig:proposed_method}(a).
In MaskGIT, the generation is started with `Masked low resolution codes' which is filled with the [MASK] code, the
VQGAN code which indicates that it is masked, for all locations.
Then, the low resolution model predicts the probabilities for all the [MASK] locations in parallel,
and samples a VQGAN code based on its predicted probabilities over all possible VQGAN codes for each location.
The location with the low probability is replaced with the [MASK] code again, and the VQGAN codes are resampled by
predicting the probabilities using the low resolution model.
This process is repeated in $T$ steps.
At each iteration, the model predicts all VQGAN code simultaneously but only keeps the most confident ones.
The remaining VQGAN codes are replaced with the [MASK] code and re-predicted in the next iteration.
The mask ratio during the iterations is determined by $\cos\left(\frac{\pi}{2} \frac{t}{T}\right)$, where $t$ indicates the current iterations in the
total steps $T$.
This mask ratio is monotonically decreasing from 1 to 0 with respect to $t$, which ensures that most of the locations are
masked during the early stage in the iterations to prevent producing the inconsistent codes.

At the second stage, the high-resolution codes are generated using the high resolution model, where the generation
process is almost same as the low resolution model.
The difference from the low-resolution model is that the model accepts the low-resolution image generated at the first
stage as an additional conditional image, and generates NFoV images, as shown in Fig.~\ref{fig:proposed_method}(a).
To integrate the generated NFoV images into an omni-directional image, the overlapped regions are merged with weights
depending on the distance from the centers of the NFoV images.
Specifically, let $x_i$ and $x_j$ be the two overlapping pixel values, and let $d_i$ and $d_j$ be the distances from the
centers of the NFoV images at each position.
The integrated pixel value $y$ is given by

\begin{equation}
    y = \frac{w_i}{(w_i + w_j)}x_i + \frac{w_j}{(w_i + w_j)}x_j
\end{equation}
where \( w_i = 1 - \frac{di}{\max_k(d_k)} \), \( w_j = 1 - \frac{dj}{\max_k(d_k)} \).

The network architecture used in the low-resolution model and the high resolution model is shown in Fig.~\ref{fig:attention_architecture}.
The layer structure was adapted from MaxViT~\cite{maxvit}, although Transformer has been used in MaskGIT~\cite{maskgit}
and Muse~\cite{muse}.
The 8-layer MaxViT models were used in both the low-resolution model and the high-resolution model.
In the low-resolution model, the padding in MBConv was replaced to the circular padding to encourage the continuity at the
edges in ERP, whereas it was remained to the zero padding in the high-resolution model.
The block attention was applied within each divided region at the low-resolution model and within each NFoV image in
the high-resolution model, as shown in Fig.~\ref{fig:attention_architecture}.
The grid attention was also applied globally in sparse at the low-resolution model as in the original MaxViT model,
whereas it was applied among same locations over the NFoV images at the high-resolution model.
\subsection{Training}\label{subsec:training}
\begin{figure}[t]
    \centering
    \includegraphics[width=\linewidth]{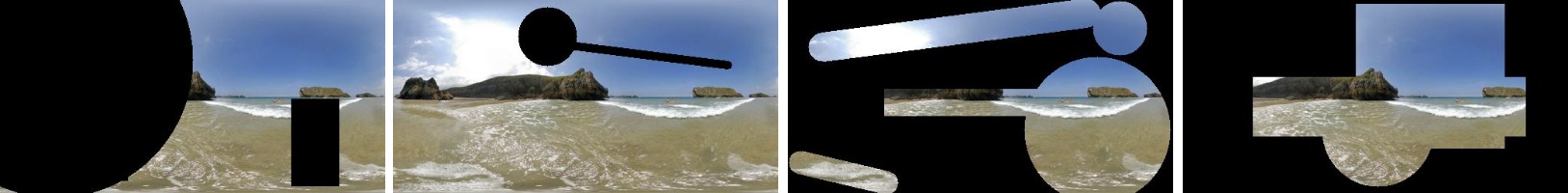}
    \caption{
        Examples of conditional image in training.
        These images are generated from omni-directional images in ERP by randomly masking.
    }
    \label{fig:mask_example}
\end{figure}

The low-resolution and high-resolution models are independently trained, as shown in Fig.~\ref{fig:proposed_method}(b).
The objective of the training is to make the low-resolution and high-resolution models predict plausible VQGAN codes
at [MASK]-code locations for each inference step.
For the low-resolution model, the inputs are `randomly masked low-resolution codes' and a conditional image which
emulates the NFoV image embedded at the center in ERP (Fig.~\ref{fig:proposed_method} a).
The mask ratio in the randomly masked low-resolution codes is set to $\cos\left(\frac{\pi}{2} r\right)$, where $r$ is sampled from a uniform
distribution $[0,1)$, since this emulates the single iteration in the inference.
Examples of the conditional image in the training are shown in Fig.~\ref{fig:mask_example}.
They are prepared by randomly masking an original omni-directional image in ERP to emulate the conditional image in the
inference.
Since they are not limited to the single NFoV image embedded in ERP, the trained model can be applied to various
in-painting and out-painting tasks, as described in~\ref{subsec:evaluation-of-input-images-under-different-conditions}.
The low-resolution model is trained to predict the original VQGAN codes in the real omni-directional image at
[MASK]-code locations, so that the cross entropy is used as the loss function to train the model.

For the high-resolution model, the inputs are `randomly masked high-resolution codes' for multiple NFoV images
(26 NFoV images in our implementation), conditional NFoV images converted from the conditional image in ERP,
and the reconstructed low-resolution NFoV images converted from the low-resolution omni-directional image reconstructed
using the pre-trained VQGAN encoder and decoder.
The `randomly masked high-resolution code' and the conditional image in ERP are prepared in the same manner for the
low-resolution model.
The low-resolution omni-directional image is required in the inputs of the high-resolution model to emulate the
low-resolution image generated at the first stage.
The high-resolution model is trained to predict the original VQGAN codes in the NFoV images converted from the real
omni-directional image at [MASK]-code locations based on the cross-entropy loss function.

Although the first and second stages are sequentially processed in the inference, they are independently trained in
parallel, as shown in Fig.~\ref{fig:proposed_method}(b).
This property is advantageous to shorten the required training time if multiple GPUs
(Graphics Processing Units) can be used, although a single GPU was used in our implementation.

%% file: sec/4_experiment.tex
\section{Experiments}
The omni-directional image dataset, SUN360~\cite{sun360}, was used in the experiments.
The 5,000 outdoor images were used for test, while the remaining 47,938 outdoor images were used for training.
The size of the images in the dataset is 512$\times$1024 pixels.
Although the proposed method generates images in 1024$\times$2048 pixels, it was resized to 512$\times$1024 pixels for the evaluation.
For comparison, several conventional methods were also evaluated, including GAN-based methods with CNN (Convolutional Neural Networks)~\cite{sphericalgan} or MLP-Mixer~\cite{odimlpmixergan}
, a VQGAN-based method (OmniDreamer)~\cite{omnidreamer}, and a GAN-based in-painting method (LAMA)~\cite{lama}.
The models were implemented in PyTorch, and were trained in a single GPU (NVIDIA RTX3090).
The code for the network architecture in the proposed method is provided in the supplementary material.

The pre-trained VQGAN was obtained from~\cite{taming-transformers-pytorch}, which is the model with 1024 codebooks trained on ImageNet.
Eight MaxViT layers in Fig.~\ref{fig:attention_architecture} were used in both the low-resolution and high-resolution models, with 256 internal dimensions.
The sizes of the VQGAN codes at the first and second stages were 16$\times$32 patches and 16$\times$16 patches $\times$26 NFoV images, respectively.
These VQGAN codes were converted into trainable feature vectors in the 256 dimensions.
The conditional image was also down-sampled into the same size as the VQGAN codes with 256 dimensions using CNN.
At each iteration of MaskGIT in the first stage, these two feature vectors were added with trainable positional encoding vectors, and then were inputted into the low-resolution model composed of the 8 MaxViT layers.
At each iteration in the second stage, the low-resolution image generated at the first stage (inference) or reconstructed using the pre-trained VQGAN (training) was converted into NFoV images.
These NFoV images were down-sampled with CNN to be added with the feature vectors of the input VQGAN codes, the conditional NFoV images, and the positional encoding, and then were inputted into the high-resolution model.
The total steps $T$ in MaskGIT was set to 16 at both stages.

The optimizer for training the models was AdamW with the learning rate of 0.001, the weight decay of 1e-5, Amsgrad,
and the learning-rate scheduling of ExponentialLR reducing it by 0.95 every 5,000 iterations.
OmniDreamer~\cite{omnidreamer} was trained for 14 days (30 epochs in all training stages), while the proposed method
was trained over 4 days (2 days with the batch size 16 at the first stage, and 2 days with the batch size 8 in the second stage,
for 180,000 iterations at each stage).
The other conventional methods were trained for 4 days, using their default batch sizes and hyper-parameters.
FID (Frechet Inception Distance)~\cite{fid}, IS (Inception Score)~\cite{is}, and LPIPS (Learned Perceptual Image Patch Similarity)~\cite{lpips} were used for evaluating the synthesized omni-directional images in ERP.

%% file: sec/5_result.tex
\section{Results}

\subsection{Evaluation of Proposed Method}
\begin{table}[tbp]
    \centering
    \caption{Quantitative comparision with conventional methods}
    \begin{tabular}{lccc}
        \toprule
        Method & IS (↑) & FID (↓) & LPIPS (↑) \\
        \midrule
        2S-ODIS (Proposed method) & \textbf{5.969} & \textbf{18.263} & 0.662 \\
        2S-ODIS (2days) & 5.857 & 18.656 & \textbf{0.668} \\
        OmniDreamer~\cite{omnidreamer} & 4.458 & 23.101 & 0.655 \\
        CNN-based cGAN~\cite{sphericalgan} & 4.684 & 40.049 & 0.633 \\
        MLPMixer-based cGAN~\cite{odimlpmixergan} & 4.402 & 47.690 & 0.634 \\
        LAMA~\cite{lama} & 5.784 & 69.485 & 0.478 \\
        \bottomrule
    \end{tabular}
    \label{tab:quantity_eval}
\end{table}
\begin{figure}[tbp]
    \centering
    \includegraphics[width=\linewidth]{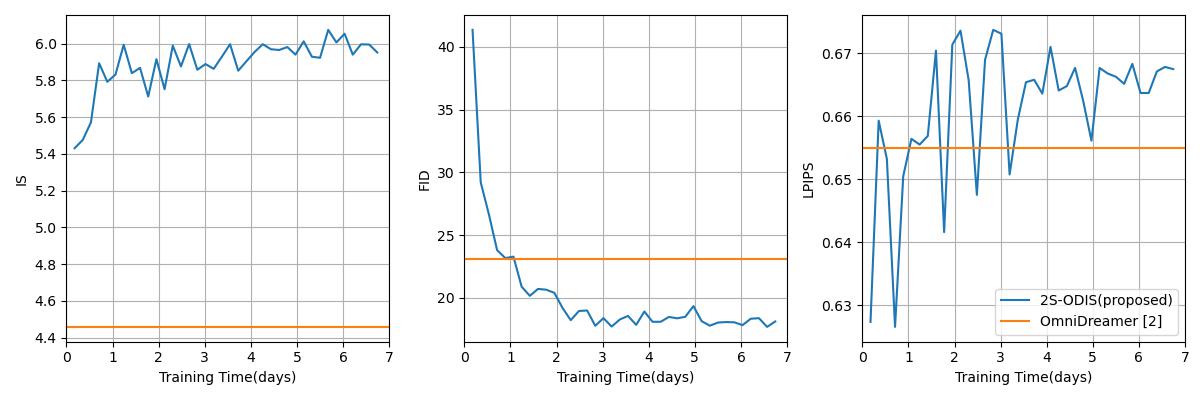}
    \caption{Evaluation metrics during training of proposed method compared with conventional method, OmniDreamer~\cite{omnidreamer}}
    \label{fig:quantity_graph}
\end{figure}
\begin{figure}[tbp]
    \centering
    \includegraphics[width=1\linewidth]{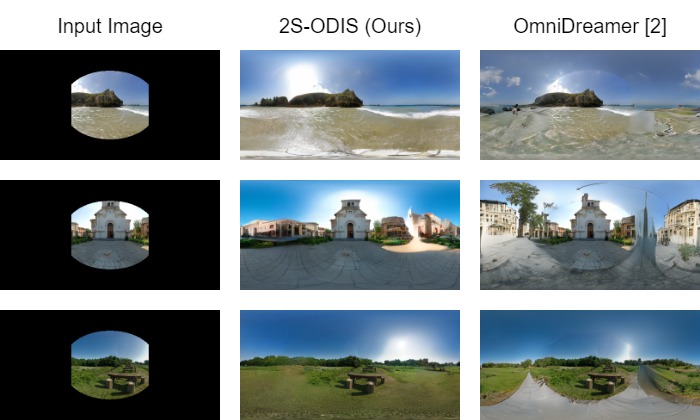}
    \caption{Examples of synthesized omni-directional images compared with conventional method, OmniDreamer~\cite{omnidreamer}}
    \label{fig:quality_evaluation}
\end{figure}
\begin{figure}[tbp]
    \centering
    \includegraphics[width=0.6\linewidth]{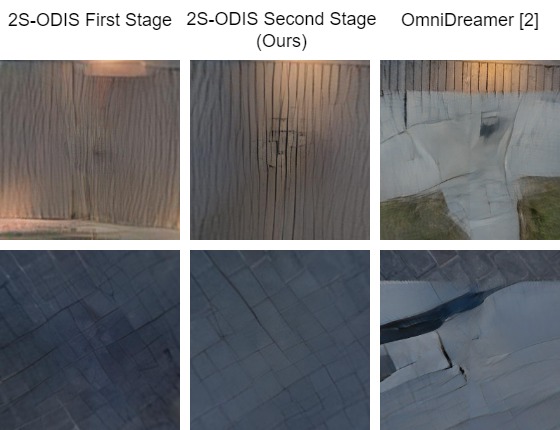}
    \caption{Examples of NFoV images toward ground extracted from synthesized omni-directional images in proposed and conventional methods}
    \label{fig:distortion}
\end{figure}
The proposed method, 2S-ODIS, was quantitatively evaluated, compared with the conventional methods,
OmniDreamer~\cite{omnidreamer}, CNN-based cGAN~\cite{sphericalgan}, MLPMixer-based cGAN~\cite{odimlpmixergan},
 and LAMA~\cite{lama}, as shown in Table~\ref{tab:quantity_eval}.
 The models in the proposed method was trained for 4 days (2 days for low-resolution model and 2 days for high-resolution model).
 For comparison, the result with the models trained for 2 days (1 day + 1 day for low-resolution and high-resolution models) was also shown in the table.
The field of view for a NFoV image embedded in an input conditional image was set to 126.87 and 112.62 degrees for width and height in the experiments, respectively.
It can be seen from the table that the proposed method achieved higher performance than the other conventional methods.
 Although the highest performance was achieved in the proposed method trained for 4 days,
 the performance already outperformed the other methods even with the 2-day training.
 To see this more clearly, the evaluation metrics during the training of the proposed method is shown in Fig.~\ref{fig:quantity_graph},
 where the performance exceeded OmniDreamer by 2 days and converged in around 4 days.
 Thus, the proposed method drastically shortened the training to 2-4 days from 14 days in OmniDreamer including the fine-tuning of the VQGAN model.
 In addition, the inference in the proposed method was much faster than in OmniDreamer: 1.54 seconds and 39.33 seconds for synthesizing each omni-directional image in the proposed method and OmniDreamer, respectively.
 This is because the proposed method is based on the simultaneous VQGAN-code prediction instead of the sequential auto-regressive prediction.

For the qualitative comparison, the examples of the synthesized omni-directional images are shown in Fig.~\ref{fig:quality_evaluation}.
 It is clear that the proposed method generated globally plausible and locally detained omni-directional images,
 while OmniDreamer~\cite{omnidreamer} sometimes failed in generating continuous images, especially along the edges in the input conditional images.
 To see if the proposed method can generate NFoV images without the distortion,
the NFoV images toward the ground were extracted from the synthesized omni-directional images,
since the geometric distortion is large at the poles in ERP.
 The examples of the NFoV images toward the ground are shown in Fig.~\ref{fig:distortion}.
 The left column shows the NFoV images at the first stage, while the middle column shows the NFoV images at the second stage.
 Since the pre-trained VQGAN code cannot represent the geometric distortion in ERP, the straight lines were not appropriately reproduced at the first stage.
 However, they were compensated at the second stage, which can be clearly seen in the sample images.
 On the other hand, OmniDreamer cannot appropriately reproduce the NFoV images toward the ground.
 Thus, it was confirmed that the proposed method synthesized omni-directional images without geometric distortion.

\subsection{Evaluation in Various Input Conditions}\label{subsec:evaluation-of-input-images-under-different-conditions}
\begin{table}[tbp]
    \centering
    \caption{Quantitative comparison in various conditional images}
    \scalebox{0.9}{
        \begin{tabular}{clccc}
            \toprule
            Mask Setting & Method & IS (↑) & FID (↓) & LPIPS (↑) \\
            \midrule
            Inpainting & 2S-ODIS (Proposed) & \textbf{5.582} & \textbf{15.044} & 0.685 \\
            & OmniDreamer~\cite{omnidreamer} & 4.672 & 41.209 & \textbf{0.708} \\
            \midrule
            Inpainting of & 2S-ODIS (Proposed) & \textbf{6.084} & \textbf{13.038} & 0.680 \\
            Ground Region & OmniDreamer~\cite{omnidreamer} & 5.474 & 15.303 & \textbf{0.699} \\
            \midrule
            Outpainting from & 2S-ODIS (Proposed) & \textbf{5.722} & \textbf{19.437} & 0.663 \\
            Two Images & OmniDreamer~\cite{omnidreamer} & 3.952 & 33.403 & \textbf{0.672} \\
            \bottomrule
        \end{tabular}
    }
    \label{tab:other_task}
\end{table}
\begin{figure*}[tbp]
    \centering
    \includegraphics[width=\linewidth]{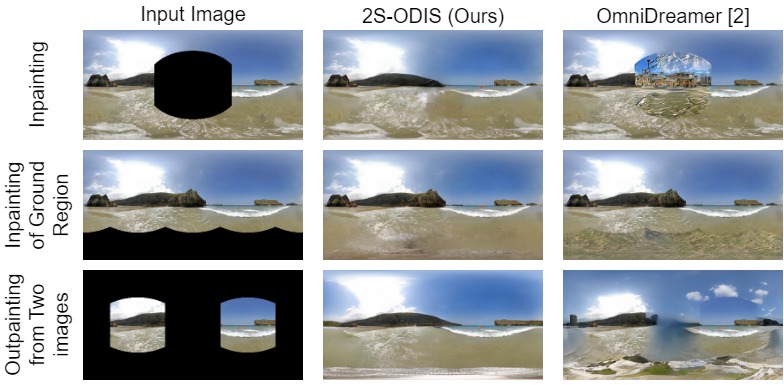}
    \caption{Examples of synthesized omni-directional images in various input conditions.}
    \label{fig:abl_quality_evaluation}
\end{figure*}
As explained in~\ref{subsec:training}, the models in the proposed method were trained with various conditional images in Fig.~\ref{fig:mask_example}.
They were not limited to the single NFoV image embedded in ERP,
so that the proposed model can be applied various in-painting and out-painting tasks.
For example, the model can be applied to the in-painting task to remove objects or people in the omni-directional image
taken by a 360-degree camera, as shown in the top row in Fig.~\ref{fig:abl_quality_evaluation}.
Another example is the task to fill in the ground region of an omni-directional image as shown in the middle row in Fig.~\ref{fig:abl_quality_evaluation},
since the omni-directional image often includes a hand or a camera stand at the bottom region in ERP.
The last example in Fig.~\ref{fig:abl_quality_evaluation} is to synthesize an omni-directional image which includes
two NFoV images such as front and rear cameras of a smartphone.
Although OmniDreamer failed in synthesizing the omni-directional images in these situations, the proposed method
generated high-quality omni-directional images.
The quantitative results shown in Table~\ref{tab:other_task} also indicate that the proposed method achieved higher performance than OmniDreamer.
Although the diversity of the synthesized images was higher in OmniDreamer than the proposed method,
it may be due to generating random images as shown in Fig.~\ref{fig:abl_quality_evaluation}.

\subsection{Ablation Study}
    \begin{table}[tbp]
        \centering
        \caption{Ablation study in propose method}
        \begin{tabular}{clccc}
            \toprule
            & & IS (↑) & FID (↓) & LPIPS (↑) \\
            \midrule
            (1) & Proposed & \textbf{5.969} & \textbf{18.263} & 0.662 \\
            (2) & 1 Stage: Low Resolution Model & 5.798 & 28.329 & \textbf{0.670} \\
            (3) & 1 Stage: High Resolution Model & 4.821 & 52.453 & 0.638 \\
            (4) & Direct use of low-resolution VQGAN codes & 5.837 & 21.820 & 0.663 \\
            \bottomrule
        \end{tabular}
        \label{tab:ablation_study}

    \end{table}

 Ablation studies were conducted to investigate the effectiveness of each component in the proposed method:
 the low-resolution and high-resolution models in the 2-stage structure.
 The results are shown in Table~\ref{tab:ablation_study}.
 (1) is the proposed method with the 2-stage structure, while (2) and (3) are the results with 1-stage structure only
using the low-resolution model and the high-resolution model, respectively.
 As can be seen from the table, the 2-stage structure was indispensable for the high-quality image synthesis.

 Moreover, it was examined that the low-resolution VQGAN codes generated at the first stage were directly used at the
second stage instead of the low-resolution image generated at the first stage, since the 2-stage structure in Muse~\cite{muse}
uses the low-resolution VQGAN codes directly.
 The result is shown in Table~\ref{tab:ablation_study} (4).
 It was confirmed from the result that the low-resolution image was better to use at the second stage than the low-resolution VQGAN codes directly.
 This may be because the low-resolution image is compressed by CNN to properly extract the global information generated at the first stage.

%% file: sec/6_limit.tex
\section{Limitations and Future Prospects}

The proposed method synthesizes an omni-directional image by merging multiple NFoV images with weights depending on the distance from the edges of the NFoV images.
 Although the generation of the NFoV images are conditioned by the global low-resolution image generated at the first stage, it may be possible to generate discontinuous NFoV images.
 One possible solution would be to add an additional network to refine the generated omni-directional images to improve the continuity between NFoV images.
 Another issue is that it takes 1-2 days to convert omni-directional images in the dataset to VQGAN codes using the pre-trained VQGAN encoder.
 This may be alleviated by constructing a light-weight encoder using model distillation of the VQGAN encoder.

Currently, inputs in the proposed method were limited to the conditional images such as a single or several NFoV images embedded in ERP, or masked omni-directional images for in-painting.
 However, the proposed architecture can be applied to any conditional information such as text information, a class label,
and guidance information for style modulation using an additional module such as cross attention similar to stable diffusion.
 In addition, the hyper-parameters have not been thoroughly explored in the evaluation of the proposed method.
 For example, the directions of NFoV images at the second stage were fixed to 26 directions corresponding to the faces in rhombicuboctahedron, in addition to the field of view fixed to 60 degrees, which may be optimized in the future work.
 Furthermore, the network structure such as MaxViT might be improved to more optimized architecture to the omni-directional image synthesis.
 Although this paper focused on the tasks of omni-directional image synthesis, the proposed architecture would be useful for other omni-directional image tasks, such as semantic segmentation and object detection.

%% file: sec/7_conc.tex
\section{Conclusion}
A novel method for the omni-directional image synthesis is proposed in this paper.
 By using the pre-trained VQGAN encoder and decoder without fine-tuning, the training of the model was drastically shortened.
 To manage the distortion in an omni-directional image in ERP, a two-stage structure was adopted.
 At the first stage, an omni-directional image was generated in ERP without geometric distortion correction, so that it cannot reproduce straight lines at poles in a sphere.
 Therefore, it was corrected at the second stage by synthesizing an omni-directional image from multiple NFoV images based on geometric distortion correction.
 To realize fast inference, the sampling strategy in MaskGIT was adopted to predict VQGAN codes simultaneously.
 As a result, the proposed method achieved the high-quality omni-directional image synthesis with low computational costs both in training and inference.

%% file: sec/X_suppl.tex
\section{Source codes of network architecture}
The source codes for the network architecture in the proposed method are shown in Program~\ref{maxvit_code} and Program~\ref{maxvit_layer_code}.
Program~\ref{maxvit_code} is the source code of the low-resolution model at the first stage, where MultiAxisTransformerLayer is the
MaxViT layer provided in Program~\ref{maxvit_layer_code}.
The input $x$ is added with the positional encoding and the feature map of the conditional image compressed by CNN.
The source code of the high-resolution model is almost same as Program~\ref{maxvit_code}, except that the feature map of the NFoV image
extracted from the low-resolution image at the first stage is additionally added to the input $x$.
\begin{figure}[b]%[tbp]
    \begin{lstlisting}[language=Python,caption=Source code of MaxViT at first stage in PyTorch,label=maxvit_code]
class MultiAxisTransformer(nn.Module):
    def __init__(self, vocab_size, d_model,layer=8,seq_len=512):
        # vocab_size: The number of VQGAN Codebooks
        # d_model: The number of dimentions of model
        # layer: The number of MaxViTLayers
        # seq_len: The resolution of the input latent variables
        super(MultiAxisTransformerModel, self).__init__()
        self.positional_encoding = nn.Parameter(torch.randn(1,seq_len,d_model))
        self.embedding = nn.Embedding(vocab_size+1, d_model)
        self.trans_layers = nn.ModuleList([MultiAxisTransformerLayer(d_model,8) for i in range(layer)])
        self.mask_condition_conv = CNN(d_model)
        self.fc = nn.Linear(d_model, vocab_size)

    def forward(self, x, masked_condition,*args):
        x = self.embedding(x)
        condition = einops.rearrange(self.mask_condition_conv(masked_condition),"b c h w->b (h w) c")
        x = x + self.positional_encoding + condition
        for layer in self.trans_layers:
            x = layer(x)
        output = self.fc(x)
        return output
    \end{lstlisting}
\end{figure}
\begin{figure}[tbp]
\begin{lstlisting}[language=Python,caption=Source code of MaxViT layer in PyTorch,label=maxvit_layer_code]
class MultiAxisTransformerLayer(nn.Module):
    def __init__(self, d_model, n_head, patch_size=4):
        # d_model: The number of dimentions of model
        # n_head: Number of attention heads
        # patch_size: The number of patch size of block attention

        super().__init__()
        self.mbconv = MBConvBlock(d_model)
        self.block_attention = nn.TransformerEncoderLayer(d_model, n_head,dim_feedforward=d_model*4, batch_first=True,norm_first=True)
        self.grid_attention = nn.TransformerEncoderLayer(d_model, n_head,dim_feedforward=d_model*4, batch_first=True,norm_first=True)
        self.patch_size = patch_size

    def forward(self,x):
        x = einops.rearrange(x, "b (h w) c->b c h w",h=16)
        b,c,h,w = x.shape
        x = self.mbconv(x)
        x = einops.rearrange(x,"b c (h1 h2) (w1 w2)->(b h1 w1) (h2 w2) c",
            h1=h//self.patch_size,h2=self.patch_size,
            w1=w//self.patch_size,w2=self.patch_size)
        x = self.block_attention(x)
        x = einops.rearrange(x,"(b h1 w1) (h2 w2) c->(b h2 w2) (h1 w1) c",
            h1=h//self.patch_size,h2=self.patch_size,
            w1=w//self.patch_size,w2=self.patch_size)
        x = self.grid_attention(x)
        x = einops.rearrange(x,"(b h2 w2) (h1 w1) c->b (h1 h2 w1 w2) c",
            h1=h//self.patch_size,h2=self.patch_size,
            w1=w//self.patch_size,w2=self.patch_size)
        return x
\end{lstlisting}
\end{figure}

\section{Comparison of models at first and second stages}
\begin{table}[b]%[t]
    \centering
    \caption{Comparison of models at first and second stages}
    \scalebox{1}{
        \begin{tabular}{llccc}
            \toprule
            First Stage & Second Stage & IS (↑) & FID (↓) & LPIPS (↑) \\
            \midrule
            (1) MaxViT & MaxViT & 5.969 & \textbf{18.263} & 0.662 \\
            (2) Transformer & MaxViT & 6.010 & 23.168 & 0.660 \\
            (3) MaxViT & MultiAxisTransformer~\cite{multiaxistransformer} & 5.820 & 22.274 & 0.657 \\
            (4) MaxViT & NeighborAttention (Fig.~\ref{neighboratten}) & \textbf{6.030} & 20.529 & \textbf{0.667} \\
            \bottomrule
        \end{tabular}
    }
    \label{tab:ablation_study_architecture}
\end{table}
\begin{figure}[t]
    \centering
    \includegraphics[width=0.7\linewidth]{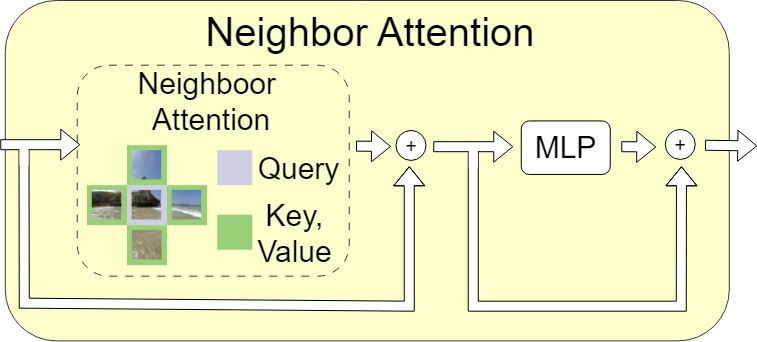}
    \caption{Neighbor attention. The attention is  calculated with adjacent NFoV images}
    \label{neighboratten}
\end{figure}

The proposed method adopted MaxViT models both at the first and second stages for the low-resolution and high-resolution models.
The different models were also examined for the models, as shown in Table~\ref{tab:ablation_study_architecture}.
(1) is the proposed method using MaxViT at both stages.
In (2), MaxViT at the first stage was replaced with Transformer.
In (3), MaxViT at the second stage was replaced with MultiAxisTransformer~\cite{multiaxistransformer}.
In (4), Grid Attention in MaxViT at the second stage was replaced with the neighbor attention in Fig.~\ref{neighboratten}.
Although IS and LPIPS were highest with (4), FID was greatly improved with (1) where MaxViT was used at both stages.
Therefore, the model (1) was adopted as the proposed method in this paper.

\section{Sample images of proposed method compared with conventional method}

\begin{sidewaysfigure}[tbp]
    \centering
    \includegraphics[width=\linewidth]{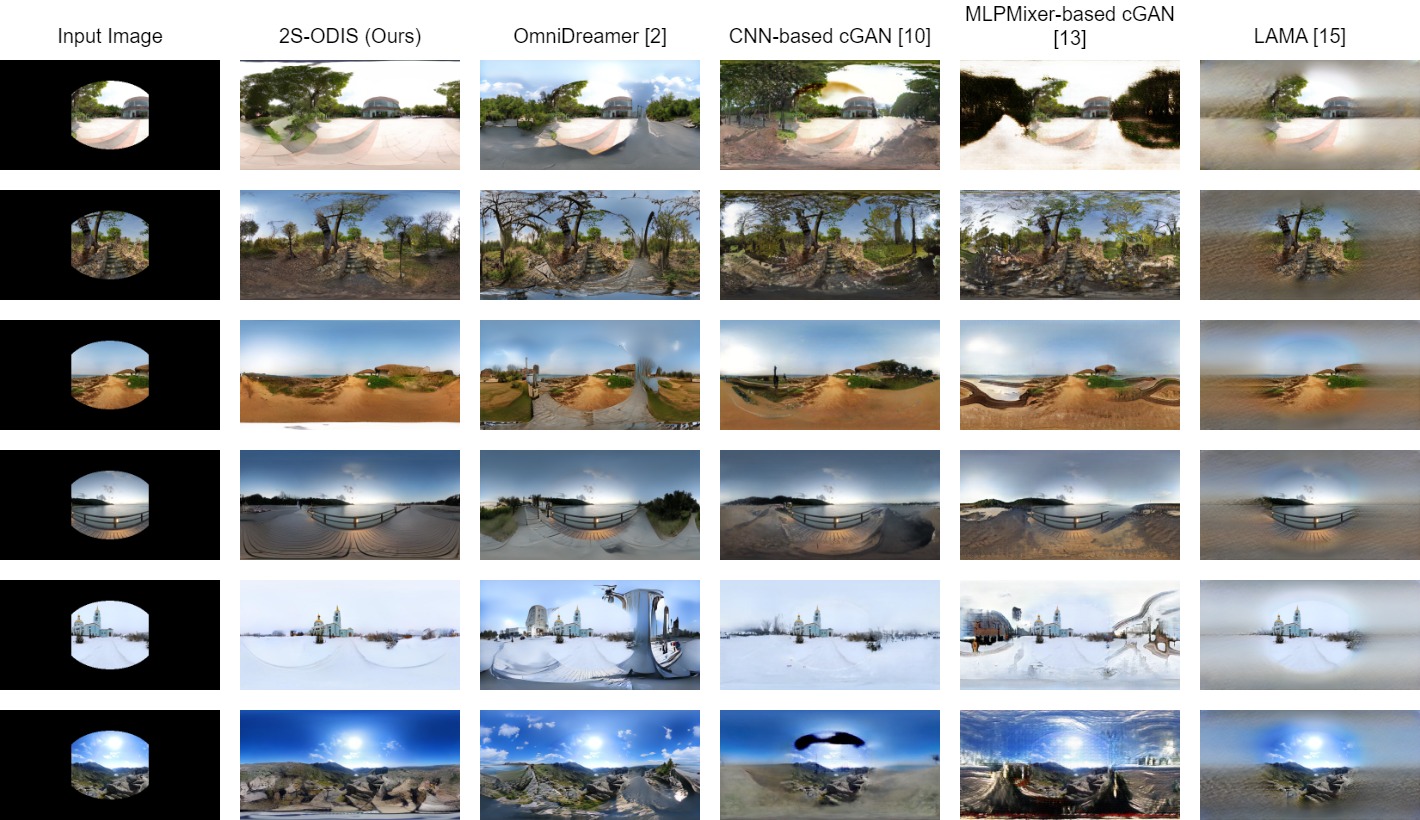}
    \caption{Additional examples of synthesized omni-directional images in Fig. 7}
    \label{fig:generated_other_sample}
\end{sidewaysfigure}
\begin{figure}[tbp]
    \centering
    \includegraphics[width=\linewidth]{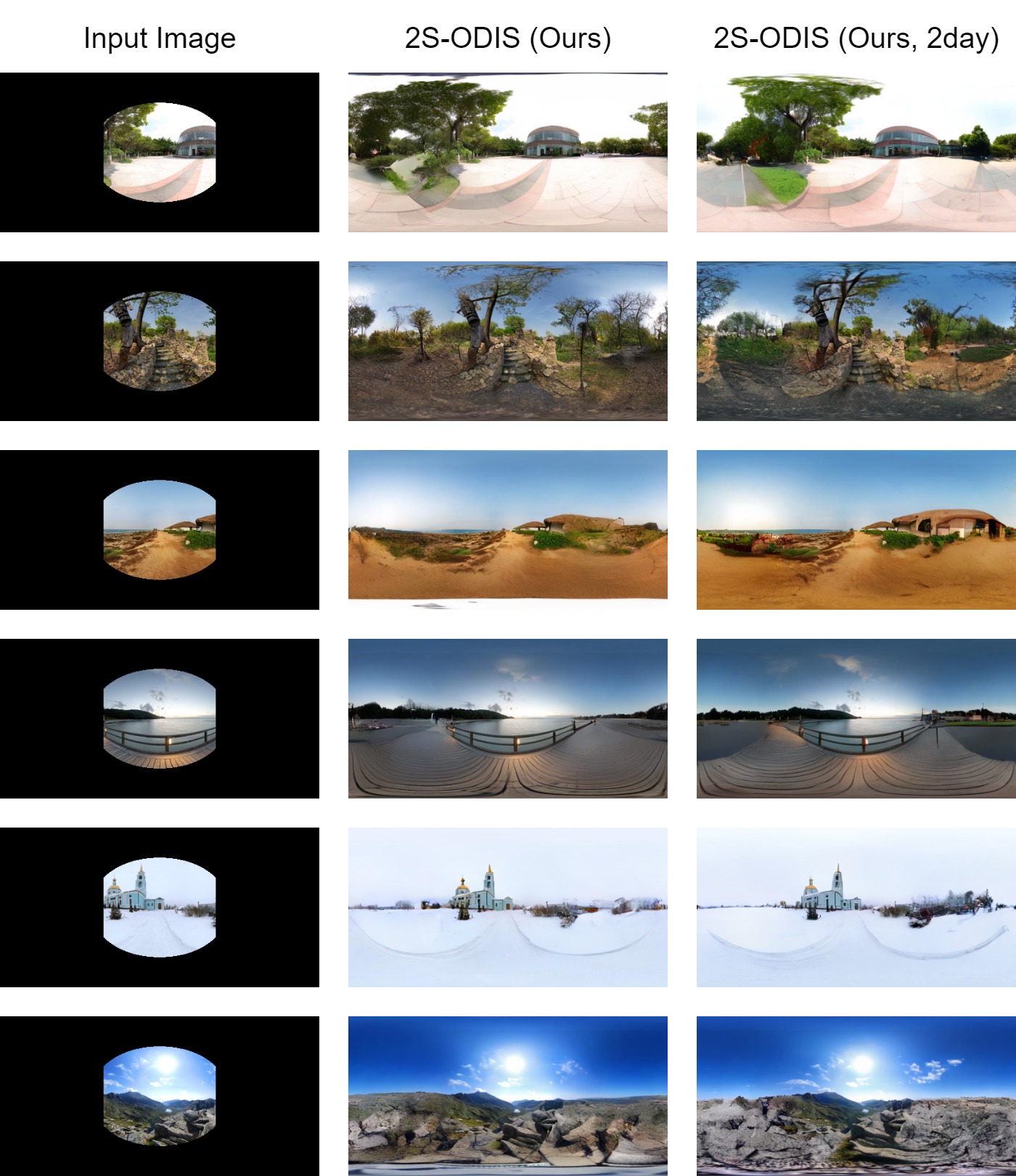}
    \caption{A comparison of synthesized omni-directional images between 2S-ODIS and 2S-ODIS 2day}
    \label{fig:compare_sample}
\end{figure}
\begin{figure}[tbp]
    \centering
    \includegraphics[width=0.9\linewidth]{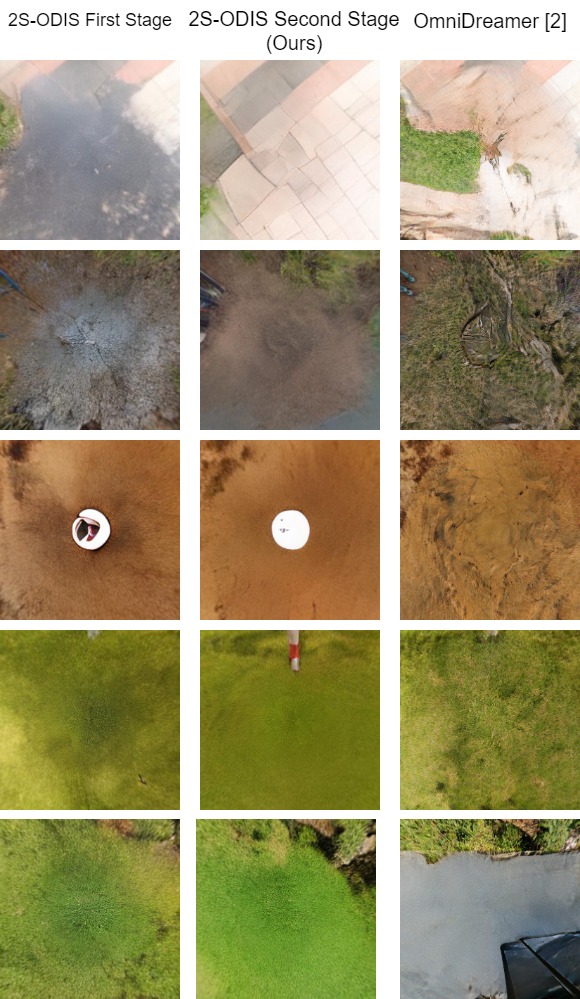}
    \caption{Additional examples of NFoV images toward ground extracted from synthesized omni-directional images in Fig. 8.}
    \label{fig:generated_other_ground_sample}
\end{figure}
Additional sample images of the synthesized omni-directional images in Fig.~7 are shown in Fig.~\ref{fig:generated_other_sample}.
A comparison of synthesized omni-directional images between 2S-ODIS and 2S-ODIS 2day are shown in Fig.~\ref{fig:compare_sample}.
Furthermore, additional sample images of the NFoV images toward the ground in Fig.~8 are shown in Fig.~\ref{fig:generated_other_ground_sample}.